\documentclass{article}

\usepackage{arxiv}

\usepackage[utf8]{inputenc} % allow utf-8 input
\usepackage[T1]{fontenc}    % use 8-bit T1 fonts
\usepackage{hyperref}       % hyperlinks
\usepackage{url}            % simple URL typesetting
\usepackage{booktabs}       % professional-quality tables
\usepackage{amsfonts}       % blackboard math symbols
\usepackage{nicefrac}       % compact symbols for 1/2, etc.
\usepackage{microtype}      % microtypography
\usepackage{lipsum}
\usepackage{graphicx}

\usepackage{times}
\usepackage{epsfig}
\usepackage{graphicx}
\usepackage{amsmath}
\usepackage{amssymb}
\graphicspath{ {./images/} }

\usepackage[linesnumbered,ruled,lined, noend]{algorithm2e}
\SetKwRepeat{Do}{do}{while}% for do while
\usepackage[flushleft]{threeparttable}
\usepackage{floatrow}
\usepackage{enumitem}
\newfloatcommand{capbtabbox}{table}[][\FBwidth]
\usepackage{soul}

\usepackage{mathtools}

\newcommand\textblue[1]{\textcolor{blue}{#1}}

\usepackage{amsmath}

\DeclareMathOperator*{\argmin}{arg\,min}

\usepackage{enumitem}

\newcommand{\comment}[1]{}

\title{MS-DARTS: Mean-Shift Based Differentiable Architecture Search}

\author{
Jun-Wei Hsieh \\
  College of Artificial Intelligence and Green Energy\\
  National Yang Ming Chiao Tung University\\
  \texttt{jwhsieh@nctu.edu.tw} \\
   \And
Ming-Ching Chang \\
  Department of Computer Science\\
  University at Albany - SUNY\\
  \texttt{mchang2@albany.edu} \\
  \And
Ping-Yang Chen \\
  Department of Computer Science\\
  National Yang Ming Chiao Tung University\\
  \texttt{pingyang.cs08g@nctu.edu.tw} \\
  \And
Santanu Santra \\
  Department of Computer Science and Engineering\\
  Yuan Ze University \\
  \texttt{santrasantanu@gmail.com} \\
  \And
Cheng-Han Chou* \\
  College of Artificial Intelligence and Green Energy\\
  National Yang Ming Chiao Tung University\\
  \texttt{aaron.cai08g@nctu.edu.tw} \\
  \And
Chih-Sheng Huang \\
   Elan Microelectronics Corp. and\\
  College of Artificial Intelligence and Green Energy\\
  National Yang Ming Chiao Tung University\\
  \texttt{chih.sheng.huang821@gmail.com} \\
}

\begin{document}
\maketitle
\begin{abstract}
Differentiable Architecture Search (DARTS) is an effective continuous relaxation-based network architecture search (NAS) method with low search cost. It has attracted significant attentions in Auto-ML research and becomes one of the most useful paradigms in NAS. Although DARTS can produce superior efficiency over traditional NAS approaches with better control of complex parameters, oftentimes it suffers from stabilization issues in producing deteriorating architectures when discretizing the continuous architecture. We observed considerable loss of validity causing dramatic decline in performance at this final discretization step of DARTS. To address this issue, we propose a Mean-Shift based DARTS (MS-DARTS) to improve stability based on sampling and perturbation. Our approach can improve bot the stability and accuracy of DARTS, by smoothing the loss landscape and sampling architecture parameters within a suitable bandwidth. We investigate the convergence of our mean-shift approach, together with the effects of bandwidth selection that affects stability and accuracy. 
%We develop two variants of MS-DARTS iterative schemes to trade-off between the search accuracy and stability. 
Evaluations performed on CIFAR-10, CIFAR-100, and ImageNet show that MS-DARTS archives higher performance over other state-of-the-art NAS methods with reduced search cost.  Code is available at \textblue{\href{https://github.com/aaron851113/MS-DARTS}{https://github.com/aaron851113/MS-DARTS}}
\end{abstract}

% keywords can be removed
%\keywords{First keyword \and Second keyword \and More}

\section{Introduction}
Recent development of convolutional neural network (CNN) architectures has advanced substantially in several fields including computer vision and language models. Manual CNN architecture design is still a common practice nowadays; However, it can take substantial amount of time and efforts. Alternatively, network architecture design process can be automated, which might lead to improved models with lower costs and fewer computational time. Neural Architecture Search (NAS) is a technique for automating this network design process, where a large set of possible architectures are explored and optimized. NAS has growing popularity that can potentially replace the manual, trial-and-error paradigm of CNN architecture design in various fields. NAS techniques ~\cite{NAS:Survey:JMLR2019,zoph2017neural,DARTS:ICLR2019} can automatically find suitable network architectures depending on the application needs. In some cases, the resulting model can outperform networks designed by human experts. This automatic architecture search is performed upon three conceptual components: 

(1) The {\em search space} defines the possible architectures as a principal representation for design optimization.
(2) The {\em search strategy} defines the exploration technique to be performed in the search space.
(3) The {\em performance estimation strategy} evaluates the predictive performance of a given architecture on unseen data. NAS approaches generally fall into two paradigms: {\em heuristic search} and {\em differentiable search}.

% --------------------------------------------
%figure Training architecture
\begin{figure*}[t]
\centerline{
\includegraphics[width=\columnwidth]{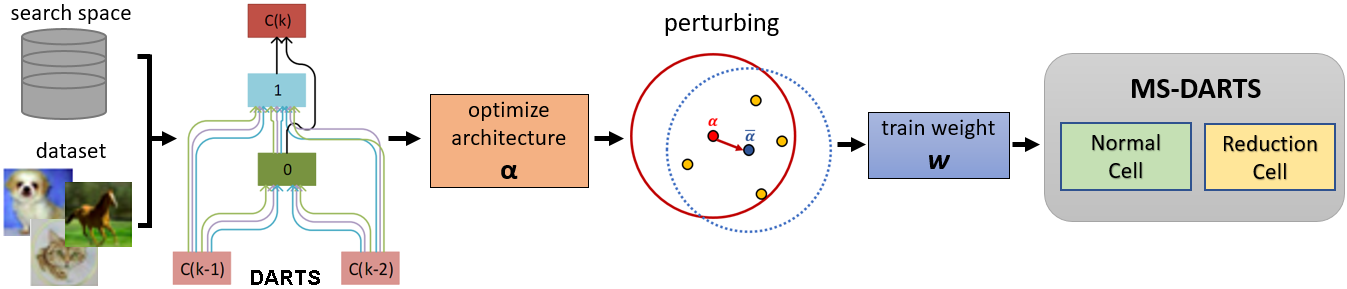}
\vspace{-0.1cm}
}
\caption{
Overview of the MS-DARTS Network Architecture Search (NAS) pipeline. The subfigure on the left is adopted from \cite{DARTS:ICLR2019}.
\vspace{-0.3cm}
}
\label{fig:training:arch:search}
\end{figure*}
% --------------------------------------------

{\bf Heuristic NAS} methods contain two sub-categories: (i) {\em evolution} based {\em e.g.}, AmoebaNet~\cite{AmoebaNet:AAAI2019} and hierarchical evolution~\cite{Hier:Evol:ICLR2018} and (ii) {\em reinforcement learning (RL)} based {\em e.g.}, NAS~\cite{zoph2017neural}, NASNet~\cite{NASNet:CVPR2018}, and ENAS~\cite{ENAS:ICML2018}. The bottleneck of heuristic search algorithms is typically the high computational cost in validating the accuracy of various architecture candidates during the model generation and optimization process. RL-based methods take accuracy as the reward to decide if a candidate model should be removed in the process. Despite their effectiveness in finding high-quality architectures, these approaches often require high computational cost (many GPU days)~\cite{zoph2017neural}. These methods are often impractical due to computational resource limitation.

{\bf Differentiable NAS} methods such as DARTS \cite{DARTS:ICLR2019,PCDARTS:ICLR2020,xie2018snas} 
build a {\em supernet} with a set of architectural parameters. Based on the supernet, the search process optimizes both network parameters and architectural parameters in a single training framework. The best architecture is generated with optimized parameters \cite{LADARTS:arXiv2020}. DARTS methods are efficient when running with limited computational resources. Despite of the computational efficiency, DARTS faces stability and generalizability issues of the obtained network architectures \cite{SantraSurvey}. Although the validation accuracy of the mixture architecture can be improved, performance of the derived architecture often collapses at the final evaluation stage, when discretizing the continuous architecture representation into a discrete one (the {\em actual}  network architecture)~\cite{zela2019understanding,Eval:Search:Phase:NAS:ICLR2020}. Such instability makes DARTS producing unwanted architectures. For example, parameter-free operations such as skip connections usually dominate the generated architecture \cite{zela2019understanding}. DARTS algorithms often prefer wide and shallow structures~\cite{Cell:NAS:ICLR2019}. In R-DARTS \cite{zela2019understanding}, early-stopping based on hand generated criteria is proposed to address these issues. However, this solution does not always work, since the intrinsic instability can occur from the beginning of the search process. The key discretization step at the end of DARTS is performed by {\em projecting} a continuous architecture onto a discrete representation manifold to derive the best discrete architecture as output. Oftentimes this projection step can cause significant performance drop between the mixture architecture (found by gradient-based optimization in super-network) and the obtained discrete architecture. It is shown in DA$^2$S \cite{DAAS:arXiv2020} that even when the super-network achieves around $90\%$ accuracy when trained on CIFAR-10 ~\cite{CIFAR-10}, sub-architecture without parameter re-training often reports less than $20\%$ accuracy on the same dataset.
DARTS adopts the Stochastic Gradient Descent (SGD) approach to optimize the cost function and then derives the desired architecture parameters. The SGD method makes DARTS mostly converge to a sharp minimum which is sensitive to perturbation. Then, slight perturbations will result insignificant cost increase and dramatically decrease the test accuracy. Such “projection gap” prevents DARTS from performing a full exploration of the architecture space in a stable way.

\textbf{Mean-Shift based DARTS.} This paper addresses this instability problem from two viewpoints, i.e., robust minimization and optimization generalization. For robust minimization, we search the optimal network architecture by minimizing its worst-case cost. For optimization generalization, we smooth out sharp minima and guide the convergence of SGD to wider and flatter minima by proposing an improved DARTS approach based on the mean-shift sampling scheme \cite{MeanShift:PAMI2002}. As shown in R-DARTS \cite{zela2019understanding}, the instability of DARTS converging into a sharp minimum is caused by the searching process mostly along a peak surface of the validation loss function. We study the stabilization of DARTS via architecture perturbation, where the mean-shift serves as our perturbation and smoothing policy that can effectively smooth out the landscape of the loss function. We are the first to investigate the adaptation of mean-shift for DARTS, and provide mathematical justification of this approach {\em w.r.t.} The ‘mean’ operation will smooth out sharp minima and guide the convergence of DARTS to flatter minima which results in good generalization and accuracy stability. Since the search surface becomes flatter, a large prediction can be made via the ‘shift’ operation so that better efficiency in architecture search is gained. The above reasons can explain why our proposed {\bf Mean-Shift based DARTS (MS-DARTS)} can stabilize and speed up the differentiable architecture search. Figure.\ref{fig:training:arch:search} overviews our approach, where mean-shift governs the projection of the continuous searched architecture into a discrete one in a flat, stable, efficient, and accurate way.

%-------------------------------

Intuitively, our optimization of the architecture search is based on perturbation and robust mean-shifting to generate flatter minima in architecture configurations, and produce benefits in optimization generation. For robust minimization, we search the optimal network architecture by optimizing its worst-case cost which is the maximum loss among its local weighted architecture candidates selected according to their performances rather than relying on a single sample. MS-DARTS essentially adopts an attention-based sampling scheme to smooth out sharp minima and thus can bypass local fluctuations to generate {\em flatter loss landscape}. It uses a Gaussian distribution to sample data points. This paper also investigates the behaviors of its bandwidth parameter to smooth out the sharp minima. We will show later that such smooth representation landscape can be quantified using the {\em eigenvalue} of the {\em Hessian matrix} of the generated architectures. Better stability can be observed by examining the accuracy differences before and after discretizing the best continuous models. Experimental results in chapter \ref{sec:experiments} show that the MS-DARTS generated architectures achieve better stability and efficiency, compared to SoTA DARTS methods without performance declining.

Contributions of this paper include:
%\vspace{-0.2cm}
\begin{itemize}[leftmargin=12pt] \itemsep -.5em
\item Proving the mean-shift design can smooth out the sharp minima caused by DARTS-based NAS to flatter minima from which significant performance drop can be avoided. 
\vspace{0.2cm}
\item Theoretical analysis is provided for the behaviors of mean-shift on affecting the stability and accuracy of DARTS, together with the effects of bandwidth selection.
\vspace{0.2cm}
\item Theoretical investigate for the effects of the bandwidth parameter of mean-shift algorithm to smooth out the sharp minima is also provided.
\vspace{0.2cm}
\item Experimental evaluations comparing MS-DARTS against state-of-the-art (SoTA) NAS models on various search spaces and image classification datasets demonstrate improvements in both accuracy and computation cost.
\end{itemize}

%%%%%%%%%%%%%%%%%%%%%%%%%%%%%%%%%%%%%%%%%%%%%%%%%
%%%%%%%%%%%%%%%%%%%%%%%%%%%%%%%%%%%%%%%%%%%%%%%%%
%%%%%%%%%%%%%%%%%%%%%%%%%%%%%%%%%%%%%%%%%%%%%%%%%

\section{Background}
Network architecture search (NAS) is a computational approach that automatically optimizes neural network architecture design. NAS is an automated method that can use limited computing resources to design the best network architecture with minimal human intervention.

\subsection{NAS}
Early NAS approaches \cite{zoph2017neural,NAS:RL:ICLR2017} train candidate architectures from scratch during each search step, thus the computation cost is very high. ENAS~\cite{ENAS:ICML2018} avoids training all candidate architectures from scratch by weight sharing. Although with speed up, this strategy may produce incorrect estimation of candidate architectures \cite{Eval:Search:Phase:NAS:ICLR2020}. It may be difficult for NAS to find a good network architecture from a large number of candidate architectures, which further reduces the effectiveness of the final searched network architecture. Subsequently, DNA~\cite{Block:DNA:CVPR2020} converts the large NAS search space into blocks to reduce parameter changes via weight sharing and can thus fully explore and train candidate architectures. DAS \cite{DAS:CVPR2018} converts the discrete network architecture search space into a continuously differentiable one, such that gradient optimization can be applied for architecture search. The primary goal of DAS is on finding hyperparameters of conv layers (filter size, number of channels, and grouped convolutions). It is observed in Maskconnect \cite{Maskconnect:ECCV2018} that cell-based network structure typically follows a pre-determined pattern between blocks, {\em e.g.}, each block only connects to its first two blocks \cite{ResNet:CVPR2016} or all previous blocks \cite{DenseNet:CVPR2017}. Similarly, \cite{CNF:NIPS2016,Budgeted:SuperNet:CVPR2018} also search for network architecture on continuous domains, where the aim is only on fine-tuning specific structures.

%%%%%%%%%%%%%%%%%%%%%%%%%%%%%%%%%%%%%%%%%%%%%%%%%
%%%%%%%%%%%%%%%%%%%%%%%%%%%%%%%%%%%%%%%%%%%%%%%%%
\subsection{DARTS}
{ \em Differentiable ARchiTecture Search (DARTS)}~\cite{DARTS:ICLR2019} methods address the above challenges by adopting a differentiable framework for architecture parameter search. Without searching over a discrete set of architectures, DARTS related works search the optimal operations in a continuous and differentiable search space, where a robust {\em cell architecture} can be efficiently determined with gradient descent. 
An important issue of DARTS is that easy-to-optimize operators (such as skip-connections and pooling operations) may dominate in early stages, hence hinder the selection of more powerful operations (such as convolutions of large kernels). 
This issue can be alleviated by freezing the updates of the architecture parameter $\boldsymbol{A}$ in the early stages, and allowing the weighting parameter W to better initialize the convolution operations \cite{PCDARTS:ICLR2020, Progressive:Diff:NAS:ICCV2019, XNAS:NIPS2019, LADARTS:arXiv2020}. In P-DARTS and DARTS+ \cite{Progressive:Diff:NAS:ICCV2019,DARTS+:arXiv2019}, a strong strategy is enforced to control the number of skip connections within a cell to a pre-determined value. The progressive search of P-DARTS \cite{Progressive:Diff:NAS:ICCV2019} gradually increases the depth of the network and reduces the candidate operations according to a mixed operation weight. This approach alleviates the problem of excessive calculations caused by the increasing of depth is alleviated and reduces search instability.

%%%%%%%%%%%%%%%%%%%%%%%%%%%%%%%%%%%%%%%%%%%%%%%%%%%
%%%%%%%%%%%%%%%%%%%%%%%%%%%%%%%%%%%%%%%%%%%%%%%%%%%
\subsection{Embedding of the evaluation procedure into the search procedure}
\label{sec:2.3}
{\em Embedding of the evaluation procedure into the search procedure} is another issue of NAS optimization, which is not explicitly performed in previous works. Various methods are designed to solve this problem ({\em e.g.}, early stopping~\cite{zela2019understanding,DARTS+:arXiv2019} and progressive optimization~\cite{Progressive:Diff:NAS:ICCV2019,SGAS:CVPR2020}), to overcome the issue of discretization gap \cite{DAAS:arXiv2020}.
It is observed in Fair-DARTS \cite{FairDARTS:ECCV2020} that the number of weak operators (such as skip connections) increases as the search proceeds, which will cause unfair competitions among the operators. Fair-DARTS \cite{FairDARTS:ECCV2020} is proposed to address this issue via relaxing the probability of operations, such that each operator has equal opportunity to develop the architecture strength. Compared to our proposed MS-DARTS, Fair-DARTS is only an indirect solution which cannot handle the discretization problem at the end. SGAS \cite{SGAS:CVPR2020} circumvents the discretization problem via a greedy strategy to prevent the problematic skip connections or other weak operators to take effect. However, potentially good operations might be pruned out as well due to this greedy under-estimation. It is frequently observed in DARTS that the resulting architecture is with good accuracy during the search stage, however performing worse in the actual testing stage.  The argues the collapse results from the unfair advantage in an exclusive competitive environment, where skip connections overly benefit from it, hence causing an aggregation. To suppress such an advantage from overshooting, they convert the competition into collaboration where each operation is independent of others. It is however an indirect approach.

%%%%%%%%%%%%%%%%%%%%%%%%%%%%%%%%%%%%%%%%%%%%%%%%%
%%%%%%%%%%%%%%%%%%%%%%%%%%%%%%%%%%%%%%%%%%%%%%%%%
\subsection{Stabilizing DARTS}
Apart from the aforementioned issues, DARTS only optimizes a single point on the simplex in each architecture search epoch. Such optimization may not generalize well after the discretization in the evaluation stage. DARTS-based algorithms prune operations on every edge except the one with the largest architecture weight. Hence the stability and generalization of DARTS has been widely challenged. 
Significant performance drop can occur in deriving the discrete architecture from the continuous version based on projection. Several approaches \cite{zela2019understanding,Random:UAI2020} are proposed to investigate this stability and generalizability issue of DARTS. 
Zela {\em et al.} \cite{zela2019understanding} empirically point out that the stability is highly correlated with the {\em dominant eigenvalue $\lambda _{\max }^{\boldsymbol{A} }$ of the Hessian matrix of the validation loss function of an architecture $\boldsymbol{A}$ }. They also present an early stopping criterion to prevent $\lambda _{\max }^{\boldsymbol{A} }$ from exploding. 
Other approaches, {\em e.g.}, partial channel connection \cite{PCDARTS:ICLR2020}, scheduled drop path \cite{NASNet:CVPR2018}, and regularization of architecture parameters are proposed to address the stability of DARTS.

%%%%%%%%%%%%%%%%%%%%%%%%%%%%%%%%%%%%%%%%%%%%%%%%%
%%%%%%%%%%%%%%%%%%%%%%%%%%%%%%%%%%%%%%%%%%%%%%%%%
%%%%%%%%%%%%%%%%%%%%%%%%%%%%%%%%%%%%%%%%%%%%%%%%%
\section{Method}
{\bf Mean-Shift based Differentiable Architecture Search.}
Our goal is to construct a novel differentiable architecture search algorithm which has a stabilizing accuracy gap between search procedure and evaluation procedure. Meanwhile, we could add less search cost as possible. We designed a perturb process with a machine learning algorithm - Mean-Shift \cite{MeanShift:PAMI2002} into DARTS search procedure, then we could balance the searching stability and search cost at the same time.
%%%%%%%%%%%%%%%%%%%%%%%%%%%%%%%%%%%%%%%%%%%%%%%%%
%%%%%%%%%%%%%%%%%%%%%%%%%%%%%%%%%%%%%%%%%%%%%%%%%
\subsection{DARTS and Mean-Shift}

\begin{figure}[b]
    \centerline{
        {\footnotesize (a)}
        \includegraphics[height=0.3\columnwidth]{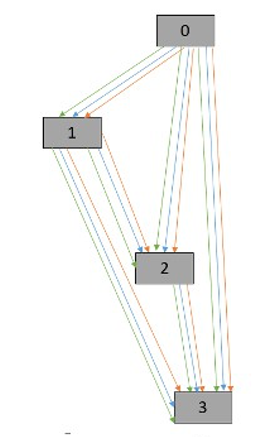}
        {\footnotesize (b)}
        \includegraphics[height=0.3\columnwidth]{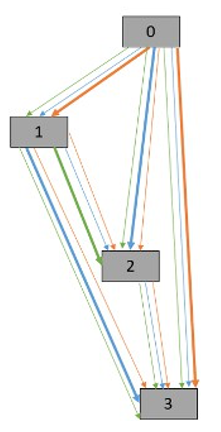}
        {\footnotesize (c)}
        \includegraphics[height=0.3\columnwidth]{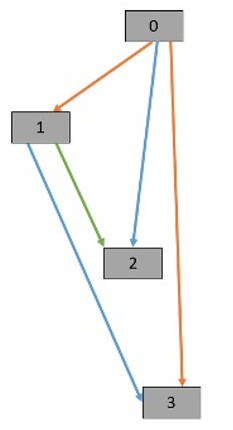}
    }
    \caption{
    DARTS \cite{DARTS:ICLR2019} training search space from continuous stage to discrete stage.}
    \label{fig:2-DARTS}
\end{figure}

{\bf DARTS}~\cite{DARTS:ICLR2019} is a cell-based neural architecture search approach. It works on a Directed Acyclic Graph (DAG) of nodes, where each node represents a set of feature maps \cite{MnasNet:CVPR2019}. Specifically, each node $x^{(i)}$ is a latent representation of feature map obtained from conv layers. 
Let ${{o}(.)}$ denotes an operation to be applied to a node $x^{(i)}$, {\em e.g.}, convolution, pooling, skip, {\em etc.}.
Each directed edge $e(i, j)$ connecting nodes $x^{(i)}$ and $x^{(j)}$ is associated with an operation $o^{(i,j)}$ that transforms node $x^{(i)}$ to node $x^{(j)}$ for $i<j$. Let ${O}$ denotes the set of all possible candidate operations.
Each intermediate node of the DAG is computed depending on all of its predecessors according to
${x}^{(j)}=\sum\limits_{i\,<j}{o^{(i,j)}({x^{(i)}})}$.
Figure.\ref{fig:training:arch:search} (left) shows the cell structure of DARTS~\cite{DARTS:ICLR2019}.As shown in Figure.\ref{fig:2-DARTS}(a), to make the search space continuous and differentiable, each operation ${o}^{(i,j)}$ transforming node $x^{(i)}$ to node $x^{(j)}$ is replaced by a ``continuous'' operation $\bar{o}^{(i,j)}$, which is obtained by mixing all possible candidate operations with SoftMax:
$
\bar{o}^{(i,j)}(x)=\sum\limits_{o\in O}{\frac{\exp (\alpha _{o}^{(i,j)})}{\sum\limits_{{o}'\in O}{\exp (\alpha _{{{o}'}}^{(i,j)})}}}o({x}),
\label{eq:softmax}
$
where $\alpha _{o}^{(i,j)}$ is a weighting parameter for  operation ${o(.)}$ from node $x^{(i)}$ to node $x^{(j)}$. 
${{\bar{o}}^{(i,j)}}(x)$ indicates a mixed result of weighted summation of outputs of all operators from node ${x^(i)}$ to node ${x^(j)}$.  
Let $W$ denote the network weights, ${\mathcal{L}_{valid}}$ and ${\mathcal{L} _{train}}$ denote the outer and inner objectives in Eq.\eqref{eq:DART:bilevel}, respectively. 
In DARTS, the searched architecture is mathematically represented as a  $d$-dimensional weighting vector $\boldsymbol{A} = \{ \alpha_o ^{(i,j)} \}$.
DARTS aims to learn the set of continuous weighting variables  $\boldsymbol{A}$ by solving the following bi-level optimization:  
%equation1
%\begin{equation}\label{eq:DART:bilevel} 
\begin{align}
\begin{split}
   & \underset{ \boldsymbol{A} }{\mathop{\min }}\,\,{\mathcal{L} _{valid}}(\boldsymbol{A} ,{{W}^{*}}( \boldsymbol{A} )) \\ 
  s.t.  & 
  { \quad { W}^{*}}( \boldsymbol{A} )=\arg \underset{W}{\mathop{\min }}\,{\mathcal{L} _{train}}( \boldsymbol{A} ,W). \\ 
  \end{split}
\label{eq:DART:bilevel}
\end{align}
%\end{equation}
%where ${\alpha}$ is the searched architecture in DARTS.  
At the end of the architecture search, a discrete architecture (Figure.\ref{fig:2-DARTS}(c)) is obtained by replacing ${{\bar{o}}^{(i,j)}}$ with the most similar operation (Figure.\ref{fig:2-DARTS}(b)): 
%equation2
%
\begin{equation}
{o^{(i,j)}}=arg \max _{o\in O} { \ \alpha_{o} ^{(i,j)}}. 
\label{eq:argmax:alpha}
\end{equation}

Despite the efficiency of DARTS, the method is not guaranteed to generalize well for the evaluation test, as the optimization is performed for a single point on the simplex in every search epoch. It is reported in \cite{chen2020stabilizing} that DARTS-based algorithm often yields deteriorating architectures, which can produce a dramatic performance drop when deriving the actual discrete architecture from the continuous mixture architecture using Eq.\eqref{eq:argmax:alpha}.

%--------------------------------------
\vspace{-0.1cm}
%--------------------------------------

%%%%%%%%%%%%%%%%%%%%%%%%%%%%%%%%%%%%%%%%%%%%%%%%%
%%%%%%%%%%%%%%%%%%%%%%%%%%%%%%%%%%%%%%%%%%%%%%%%%
{\bf Mean-shift (MS)} \cite{GradDensity:TIT1975,MeanShift:PAMI1995,MeanShift:PAMI2002}
is a well-studied non-parametric iterative algorithm for estimating the modes of probability density function (PDF) via kernel density estimation \cite{silverman1986density}. 
The principle of this algorithm is to update the mode estimation iteratively via a weighted average around neighboring points. Such weighted average estimation brings a smoothing effect to find a stationary point more stably. Let $\boldsymbol{D} \coloneqq \{\boldsymbol{A}_p\in{{\mathbb{R}}^{d}}\}_{p=1}^{N}$ denote the set of $N$  data points (or network architectures) ${\boldsymbol{A}_{p}}$, and 
$\boldsymbol{\Omega} \coloneqq \{{{\omega}_{p}}>0\}_{p=1}^{N}$ 
denote the set of weights $\omega_p$ to weight each ${\boldsymbol{A}_p}$.  In addition, let $K(\boldsymbol{A})$ be a multivariate normal kernel, {\em i.e.},
\begin{equation}
K(\boldsymbol{A}) = (2 \pi) ^{-2d} \exp \left( -\frac{\| \boldsymbol{A} \| ^2}{2}  \right),  
\label{eq:KA} 
\end{equation}
where $\| \boldsymbol{A} \| ^2$ is the norm of $\boldsymbol{A}$.  The kernel density estimate (KDE) with a kernel $K$ for $\boldsymbol{D}$ is given by:
%equation3
\begin{equation}
\hat{p}(\boldsymbol{A})=\sum\limits_{p=1}^{N} \omega_p \; K(\boldsymbol{A} - \boldsymbol{A}_p),
\label{eq:KDE}
\end{equation}
where $\int\limits_{\boldsymbol{A}}{K(\boldsymbol{A})d\boldsymbol{A}=1}$ and  $\sum\limits_{p=1}^{N}{{{\omega}_{p}}=1}$. Let $ k(z) $ be a profile, $i.e$., $k(z) = c \cdot \exp \left( -z/2  \right)$,  where $c$ is a constant and $z$ is a scalar $\in (0, \infty)$. Gaussian kernel $G(z)=-{k}'(z)$ can be regarded as a special case of $K$. The weighted mean shift $\delta_{\boldsymbol{A}}$ of the estimated density in the observation window decided by $K$ can be represented by \cite{MeanShift:PAMI2002}:
%equation4
\begin{equation}
\label{eq:meanshift} 
{\delta} _{\boldsymbol{A}} = 
\frac{
\sum\limits_{p=1}^{N} \boldsymbol{A}_p \; \omega_p \; G(\| \boldsymbol{A}-\boldsymbol{A}_{p}\| ^2 )
}{
\sum\limits_{p=1}^{N} \; \omega_p \; G( \| \boldsymbol{A}-{{\boldsymbol{A}}_{p}} \| ^2)
} - \boldsymbol{A}.
\end{equation}
Assume ${\boldsymbol{A}}^{t} $ is the solution estimated at the $t$-th iteration.   The MS algorithm updates its new estimation ${{\boldsymbol{A}}^{t+1}}$ as follows: 
\begin{equation}
\label{eq:Atplus1}
{\boldsymbol{A}}^{t+1} = 
{\boldsymbol{A}}^{t} + \delta_{\boldsymbol{A}}^{t}.
\end{equation}
The iteration continues until $\left\| {{\boldsymbol{A}}^{t+1}}-{\boldsymbol{A}}^{t}\right\| ^2 $ being convergent. We adopt this MS algorithm to tune the architecture parameters for searching a stable architecture with retained accuracy and less training time.  

%%%%%%%%%%%%%%%%%%%%%%%%%%%%%%%%%%%%%%%%%%%%%%%%%
\subsection{Motivation of Mean-Shift for DARTS}

As aforementioned, DARTS-based methods are advantageous for their fast gradient-based architecture search, however the conversion from the found continuous architecture back to discrete one can cause significant performance drop.  
Although DARTS-based methods can reduce the validation loss in the training stage, it can still be unstable in terms of numerical blow-up in computing the final architecture evaluation~\cite{zela2019understanding,DARTS+:arXiv2019}. 
Such instability is caused by the selection of SGD to train over-parameterized deep neural networks due to the problem of sharp minima \cite{Jastrzebski2018FindingFM}. Sharp minima will make the found solution highly sensitive to perturbations (or noise) and have bad generalization due to their high over-fitting to training data \cite{FlatMinima}. In DARTS the problem of sharp minima of $\mathcal{L}_{valid}$ can explain why a small perturbation $\delta$ on an architecture A will cause a significant reduction of the validation accuracy (e.g., from over 90$\%$ to less than 10 $\%$ \cite{chen2020stabilizing}). It becomes an open question whether escaping sharp minima can improve the generalization.

In this paper, the proposed MS-based DARTS is a solution to address the above issue by smoothing out the landscape of ${\mathcal{L}_{valid}}$. We compare a relevant work \cite{chen2020stabilizing} using our formulation in the following. In \cite{chen2020stabilizing}, the optimal weight $W ^* (\boldsymbol{A})$ in Eq.\eqref{eq:DART:bilevel} is re-defined to be  $\overline{W} (\boldsymbol{A} )$ as:
\begin{equation}
\overline{W} (\boldsymbol{A} ) =\argmin_{W} \; \max _{\| \delta \| \leq \epsilon } \; \mathcal{L} _{train}(\boldsymbol{A} + \delta ,W).
\label{eq:zela:alpha}
\end{equation}
The best architecture $\boldsymbol{A}^*$ can then be obtained by solving %
\begin{equation}
\boldsymbol{A}^* 
= 
\argmin_{\boldsymbol{A}} \; \mathcal{L} _{val} 
\left(
  \boldsymbol{A}, \overline{W} (\boldsymbol{A} ) 
\right).
\label{eq:delta:update:alpha}
\end{equation}
Eq.\eqref{eq:zela:alpha} solves the robust optimization problem \cite{chen2020stabilizing} minimizing the worst-case loss around a neighborhood of an architecture $\boldsymbol{A}$. In fact, at each sampled iteration, $W$ is re-estimated by changing the original architecture only once. However, this cannot provide enough a solution to flat the loss landscape of $\mathcal{L}_{train}$, which is crucial for stability control. In comparison, we propose MS-DARTS, the recursive update of the architecture $\boldsymbol{A}$ with the mean-shift algorithm in Eq.\eqref{eq:zela:alpha} before minimizing $\mathcal{L}_{train}$ can effectively smooth the landscape, such that better accuracy and stability can be obtained. One additional advantage of this recursive updating of $\boldsymbol{A}$ is efficiency, since less iterations are needed to find $W$ by minimizing $\mathcal{L}_{train}$. In addition, MS-DARTS also can address the flatness of minima. During training, flatter minima \cite{FlatMinima}, and thus better stability in architecture performance.

%%%%%%%%%%%%%%%%%%%%%%%%%%%%%%%%%%%%%%%%%%%%%%%%%
\subsection{Robust Mean-shift based DARTS}
\label{sec:MSDARTS}
The core challenge of designing robust DARTS architectures is to consider how best to increase stability while reducing the optimization gap caused by the sharp minima of $\mathcal{L}_{valid}$. In Eq.\eqref{eq:zela:alpha}, the best weight $\overline{W}$ is estimated by searching $\boldsymbol{A}$ with the worst case around a neighborhood of the original architecture. However, this optimizer cannot provide enough flatter (or smoothing) effect on the loss landscape, which is crucial for stability control. 
Our work is inspired by recent advances in understanding the loss surface of deep neural networks.  In \cite{AWL}, Izmailov {\em et al.} proposed a Stochastic Weight Averaging (SWA) scheme to smooth the loss surface by averaging the weights at different checkpoints obtained during training. In \cite{JeanCMB14}, a parameter averaging scheme was used to create ensembles in natural language processing tasks. The idea of using averaging to accelerate stochastic approximation algorithms can be traced back to the 1960s \cite{ALAS}. ‘Averaging’ forms the core idea of this paper to generate flatter minima for architecture searching by averaging multiple points along the trajectory of SGD. One additional novelty of this paper is to look ahead the direction of parameter searching during the training procedure to achieve faster convergence via a shift operation.The proposed Mean-Shift DARTS can effectively smooth the sharp landscape of $\mathcal{L}_{valid}$ by updating the architecture $\boldsymbol{A}$ using Eq.\eqref{eq:Atplus1} where $\delta_{A}$ is obtained by a boosting or filtering method. Then, Eq.\eqref{eq:zela:alpha} can be reformulated as:

\begin{equation}
  \overline{W}({\boldsymbol{A}} ) = \argmin_{W} \; \mathcal{L} _{train} ({\boldsymbol{A}}+{\delta}_{\boldsymbol{A}}, W).
  \label{eq:MSDART:opt}
\end{equation}
Based on $\overline{W} ({\boldsymbol{A}} )$, the best architecture $\boldsymbol{A}^*$ is obtained by:
\begin{equation}
\boldsymbol{A}^* = \argmin_{{\boldsymbol{A}}} 
\; 
\mathcal{L} _{valid}
\left( \boldsymbol{A}, \overline{W} ({\boldsymbol{A}} ) 
\right).
\label{eq:MSDART:Astar}
\end{equation}

%-----------------------------------

We next explain how the proposed MS-DART based on mean-shift parameter tuning can address the aforementioned instability of the found architecture $\boldsymbol{A}$ in DARTS.
Represent the architecture parameter $\boldsymbol{\alpha}$ as a $d$-dimensional vector.  
Assume there are $N$ sample points $\{ \boldsymbol{A} _p \}$ obtained around the unknown $\boldsymbol{A}$ within a radius $\epsilon$, where each 
$\boldsymbol{A}_p \in \mathbb{R}^d$.
 Different from the SWA scheme \cite{AWL}, we introduce a bandwidth parameter $h$ to control the uniform distribution between the $\epsilon$ sampling of the perturbation $\delta_{\boldsymbol{A}}$ in Eq.\eqref{eq:zela:alpha}. 
Specifically, the sequence of successive locations of the Gaussian kernel $G$ is denoted by $\{ {\boldsymbol{A}}^t \}_{t=0,1,...,T}$. Using Eq.\eqref{eq:meanshift}, we estimate $ {\delta}_{\boldsymbol{A}}^t $ as:
\begin{equation}
{\delta}_{\boldsymbol{A}}^t
=
\frac{
  \sum\limits_{p=1}^{N} \; \boldsymbol{A}^{t}_p \; \omega_p \;
    G \left( 
        \| \frac{  {\boldsymbol{A}} ^{t} - \boldsymbol{A}^{t}_p } {h} \| ^2 
      \right)
}
{
  \sum\limits_{p=1}^{N} \; \omega_p \;
    G \left( 
        \| \frac{  {\boldsymbol{A}} ^{t} - \boldsymbol{A}^{t}_p } {h}  \| ^2 
      \right)
} 
- {{\boldsymbol{A}} }^{t}, \;
t=1,2,3,...,T
\label{eq:MSDART:delta}
\end{equation}
where $\omega_p$ is used to weight ${{\boldsymbol{A}}^{t}_p}$.
Following the previous optimization works, giving higher weights to worse cases result in faster convergence and efficiently generate the adapted weights.
We also let the value of $\omega_p$ is made proportional to the validation loss of ${{\boldsymbol{A}}_p}$. This design can increase the impact of a sampling architecture ${{\boldsymbol{A}}_p}$ producing worse accuracy over other samples in our MS sampling process. This can enhance the desired landscape flattening that can improve robust architecture search. 

Given an initial architecture $\boldsymbol{A}^0$, mean-shift filtering is applied to obtain a weight, smoothed architecture $\overline{\boldsymbol{A}}$ following a few iterative steps. 
Given the initial architecture ${\boldsymbol{A}}^0$ at $t=0$, we take $N$ samples $\{\boldsymbol{A}_p \} _{p=1,...,N}$ around it within a radius $\epsilon$. Eq.\eqref{eq:MSDART:delta} is then used to compute ${\delta}_{\boldsymbol{A}}^{t=1}$ to obtain ${\boldsymbol{A}}^1$. These steps can go on iteratively until convergence. Algorithm~\ref{Algo:Algo1} describes detailed steps of this MS update scheme steps in calculating ${\overline{\boldsymbol{A}}}$.

%%%%%%%%%%%%%%%%%%%%%%%%%%%%%%%%%%%%%%%%%%%%%%%%%
\begin{algorithm}[t]
\SetAlgoLined
\textbf{Input:} Architecture ($\boldsymbol{A} , W$), bandwidth ($h$), Sampling Number ($N$), Max Iterations($T$);\\
\textbf{Output:} New Architecture ($ {\overline{\boldsymbol{A}}}$) \\ 
 Initialize $t=0$ and ${\boldsymbol{A}} ^0 = \boldsymbol{A} $;\\
\Do{ $t < T$ }{
\For{$i \in N$}{
Sample and get $ {\boldsymbol{A}} ^{t}_p $ around $ {\boldsymbol{A}} ^{t}$ within a radius $\epsilon$;\\
Get $ \omega _p $ based on the validation loss of ${\boldsymbol{A}} ^{t}_p$;
}
Compute  ${\delta}_{\boldsymbol{A}}^t$  based on Eq.\eqref{eq:MSDART:delta};\\
$ {\boldsymbol{A}}^{t+1} =  {\boldsymbol{A}}^{t} + {\delta }_{\boldsymbol{A}}^t$;\\
$t=t+1$;\\
}
\textbf{return \hspace{0.1cm} ${\overline{\boldsymbol{A}}} = {\boldsymbol{A}}^{t+1}$}
\caption{Mean-Shift Iteratively Updating Algorithm}
\label{Algo:Algo1}
\end{algorithm}
%%%%%%%%%%%%%%%%%%%%%%%%%%%%%%%%%%%%%%%%%%%%%%%%%
%------------------------------------------------

\begin{algorithm}[t]
\SetAlgoLined
\textbf{Input:} Architecture ($\boldsymbol{A}, W$), bandwidth ($h$), Sampling Number ($N$), Max Iterations ($T$);\\
\textbf{Output:} New Architecture ($\boldsymbol{A} ^*$) \\
 \While{not converged}{
  Determine $\boldsymbol{A}^*$  by minimizing ${\mathcal{L}_{val}}( \boldsymbol{A}, W )$ with Eq.\eqref{eq:MSDART:Astar}. \\
  Compute $ \overline {\boldsymbol{A}}$ by  Algorithm \ref{Algo:Algo1}\ with  parameters ( $\boldsymbol{A}^*, W$, $h$, $N$, $T$).\\
  Update $\overline{W} ({\boldsymbol{A}} )$ by minimizing ${\mathcal{L}_{train}}( \overline {\boldsymbol{A} },W )$ with Eq.\eqref{eq:MSDART:opt}. \\
 $\boldsymbol{A} = \boldsymbol{A}^* $; $W=\overline{W} (\boldsymbol{A})$.\\
 } 
 \textbf{return} Final Architecture $ (\boldsymbol{A} , W )$\\
 \caption{MS-DARTS: Updating Architecture Before Mean-shift}
\label{Algo:AMW}
\end{algorithm}
%------------------------------------------------

{\bf MS-DARTS search and update scheme.}
DARTS first search for an initial architecture $\boldsymbol{A}$. Mean-shift is then adopted to sample for architectures within the bandwidth and find out poorly-performed architectures. As discussed before, we weight more on pooly-performed architectures in consideration to better avoid trapping into bad local minimums. Eq.\eqref{eq:MSDART:opt} is then applied to update weights.
%ones produced by DARTS. 
%Mean-shift filtering can effectively reduce the performance gap between the continuous model and discrete one.  
%The sequence for architecture searching is then: optimizing Architecture $\rightarrow$ Meanshift $\rightarrow$ updating Weights.  
Algorithm \ref{Algo:AMW} describes the details of this iterative search and updating scheme for MS-DARTS.

%%%%%%%%%%%%%%%%%%%%%%%%%%%%%%%%%%%%%%%%%%%%%%%%%
\subsection{MS-DARTS Bandwidth Effects Searching Stability}
\label{sec:3.4}
The bandwidth parameter $h$ in Eq.\eqref{eq:MSDART:delta} controls the smoothing effect of the architecture search in MS-DARTS. We next discuss the effects of $h$ {\em w.r.t.} the stability of DARTS.  
We define the kernel function with a bandwidth parameter $h$ as
$
K _h (\boldsymbol{A}) 
= 
(2 \pi h ) ^{-2d} 
\exp{ (-\frac{1}{2} 
\| \frac{\boldsymbol{A}}{h} \| ^2 ) 
},
\label{eq:kernel:bandwidth}
$
and represent the DARTS loss function  $\mathcal{L}(\boldsymbol{A})$ as: 
\begin{equation}
  \mathcal{L}(\boldsymbol{A}) = \frac{1}{N} \sum\limits_{i=1}^{N}{K _h (\boldsymbol{A}-\boldsymbol{A} _p)}.
  \label{eq:loss:approx}
\end{equation}
To optimize $\mathcal{L}(\boldsymbol{A})$, let $ \tilde{\boldsymbol{A}} = \boldsymbol{A} - \boldsymbol{A}_p$, the Hessian $K_h$ is calculated as:
\begin{equation}
 \nabla ^ 2  K _h (\tilde{\boldsymbol{A}})  =  \frac{1}{h} ( \frac{1}{h} \tilde{\boldsymbol{A}}\tilde{\boldsymbol{A}} ^ T- \boldsymbol{I}) K _h (\tilde{\boldsymbol{A}}),
 \label{eq:Hessian:matrix} 
\end{equation}
where $\boldsymbol{I}$ denotes identity matrix.  The eigenvalues of  $\nabla ^ 2  K _h (\tilde{\boldsymbol{A}}) $ are:
\begin{equation}
( \frac{1}{h ^ 2}  \| {\tilde{\boldsymbol{A}} } \|^2-   \frac{1}{h} ) K _h (\tilde{\boldsymbol{A}}) 
  \text{ and }  
\frac{-1}{h} K _h (\tilde{\boldsymbol{A}}).
\label{eq:eigenvalue:bandwidth} 
\end{equation}
If $\| {\tilde{\boldsymbol{A}} } \|^2 < 2h $, the largest absolute eigenvalue is 
$ \frac{1}{h} K _h (\tilde{\boldsymbol{A}})$; otherwise, it is  
$  (\frac{1}{h ^ 2} \| {\tilde{\boldsymbol{A}}} \|^2-   \frac{1}{h} ) K _h (\tilde{\boldsymbol{A}})$.  
Note that the largest eigenvalue of $\nabla ^ 2 \mathcal{L}({\boldsymbol{A}})$ is related to the sum of eigenvalues of $\nabla ^ 2  K _h (\boldsymbol{A}-\boldsymbol{A} _p) $.
From Eq.\eqref{eq:eigenvalue:bandwidth}, the eigenvalues decrease according to the bandwidth $h$. 
The recent work \cite{zela2019understanding} shows that the performance of DARTS strongly depends on the largest absolute eigenvalue $\nabla ^ 2 \mathcal{L}(\boldsymbol{A})$.  During optimization, the smaller this largest eigenvalue, the better the DARTS performance. Our experimental results show that when a larger bandwidth $h$ is chosen, both the accuracy and convergence rate are indeed improving. This shows that the performance of MS-DARTS is highly influenced by the choice of the bandwidth $h$. 
While it is obvious that a too small bandwidth leads to noisy results, and too large bandwidth leads to over-smoothed landscapes, we empirically determine a  reasonably large bandwidth $h$ that produces the best desired smoothed landscape.
A good strategy is to track the bandwidth $h$ at each sample along the optimization trajectory, in order to find the best architecture.
We thus make a reasonable  claim that larger bandwidth $h$ leads to better smoothed landscape. 
However, too large bandwidth $h$ can over-smooth the landscape.  On the other hand, a bandwidth that is too small will lead to noisy results, where important data samples are not fully explored.  

%%%%%%%%%%%%%%%%%%%%%%%%%%%%%%%%%%%%%%%%%%%%%%%%%
%%%%%%%%%%%%%%%%%%%%%%%%%%%%%%%%%%%%%%%%%%%%%%%%%
%%%%%%%%%%%%%%%%%%%%%%%%%%%%%%%%%%%%%%%%%%%%%%%%%
\section{Experiments and Results}
\label{sec:experiments}

In this section, we first experiment our MS-DARTS across 3 search spaces on NAS-bench-1shot1 \cite{zela2020nasbench1shot1} dataset to test the searching stability via each epoch eigenvalue. Lower eigenvalue means the more stable in searching stage. Then we evaluate MS-DARTS on the architecture parameters A value to prove that our method has wider minima.
Furthermore, we evaluate MS-DARTS on CIFAR-10 \cite{CIFAR-10}, CIFAR-100 \cite{cifar100}, and ImageNet \cite{ImageNet:CVPR2009} datasets for image classification tasks, and compare its efficiency and accuracy against state-of-the-art (SoTA) NAS models. Following existing works \cite{zela2019understanding,chen2020stabilizing} on experimental setup, seven operators including skip connections were used to create 4 different search spaces. We further transfer the cells found on CIFAR-10 to CIFAR-100 and ImageNet for testing.
Finally, we test several bandwidth ($h$) values to prove our proposal in chapter \ref{sec:3.4}. The following experiments show that the suitable bandwidth ($h$) will efficiently effect the stability of MS-DARTS searching stage.

\subsection{Experiments on NAS-bench-1shot1} 
\begin{figure}[t]
\centering
    \centerline{
        \includegraphics[height=0.22\columnwidth]{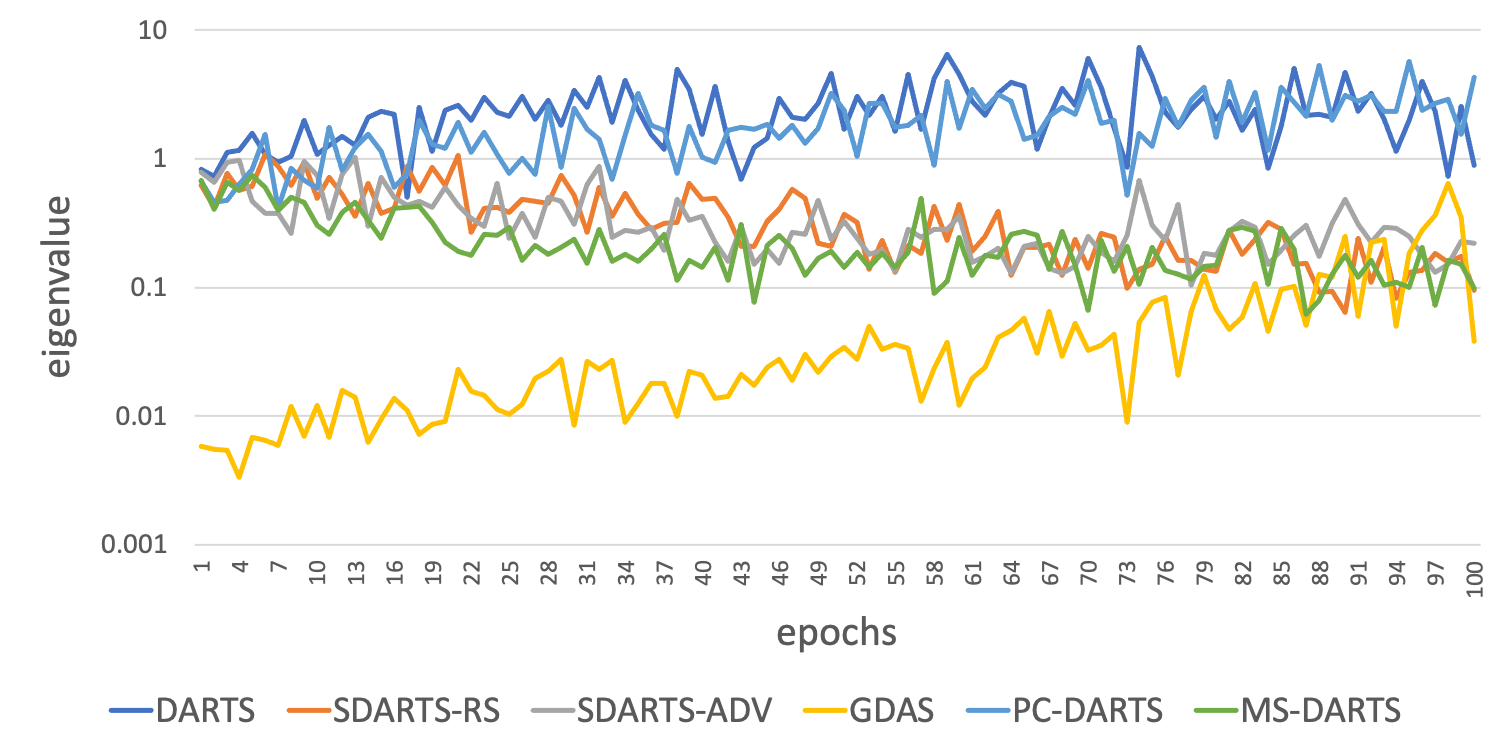}
    }
    \centerline{
        {\footnotesize (a) NAS-bench-1shot1 search space 1}
    }  
    \centerline{
        \includegraphics[height=0.22\columnwidth]{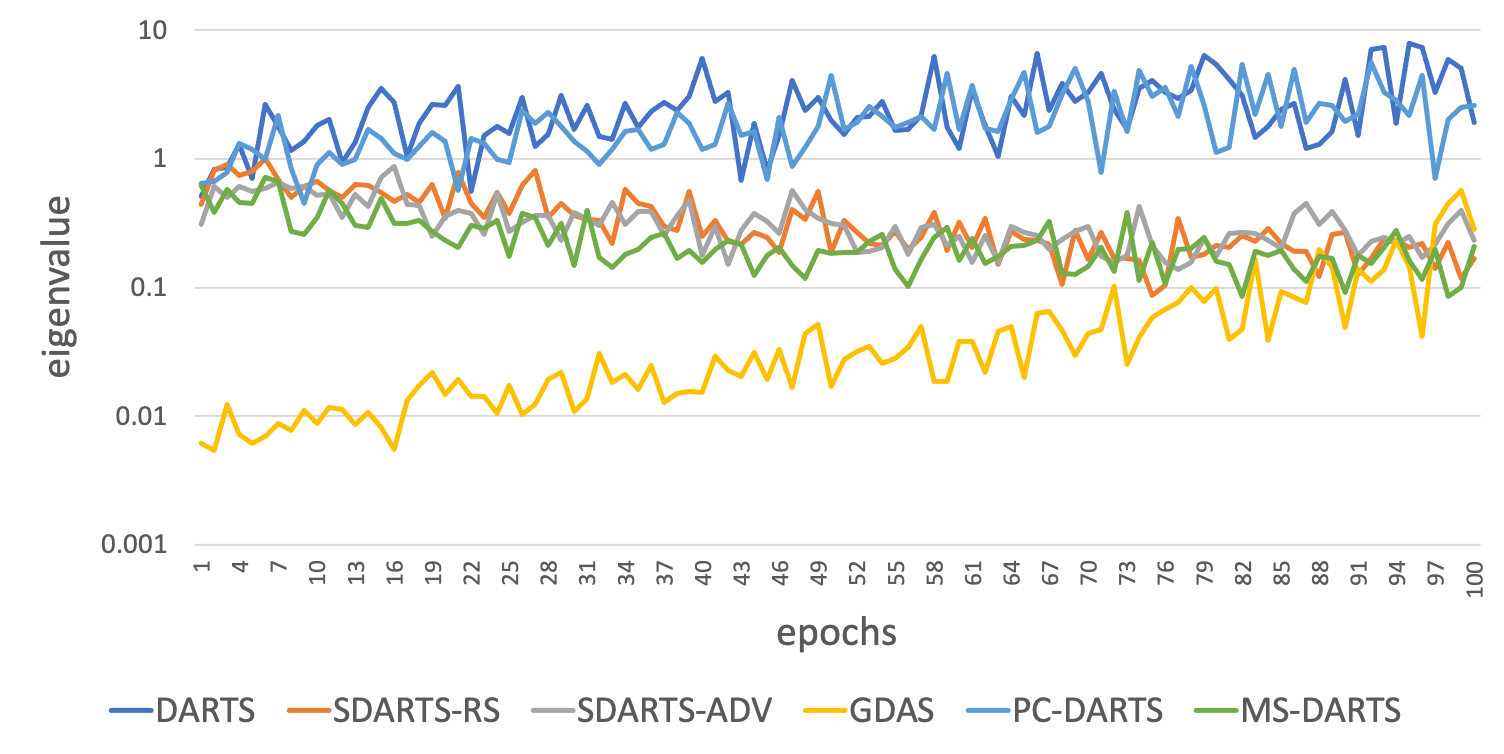}
    }
    \centerline{
        {\footnotesize (b) NAS-bench-1shot1 search space 2}
    }  
    \centerline{
        \includegraphics[height=0.22\columnwidth]{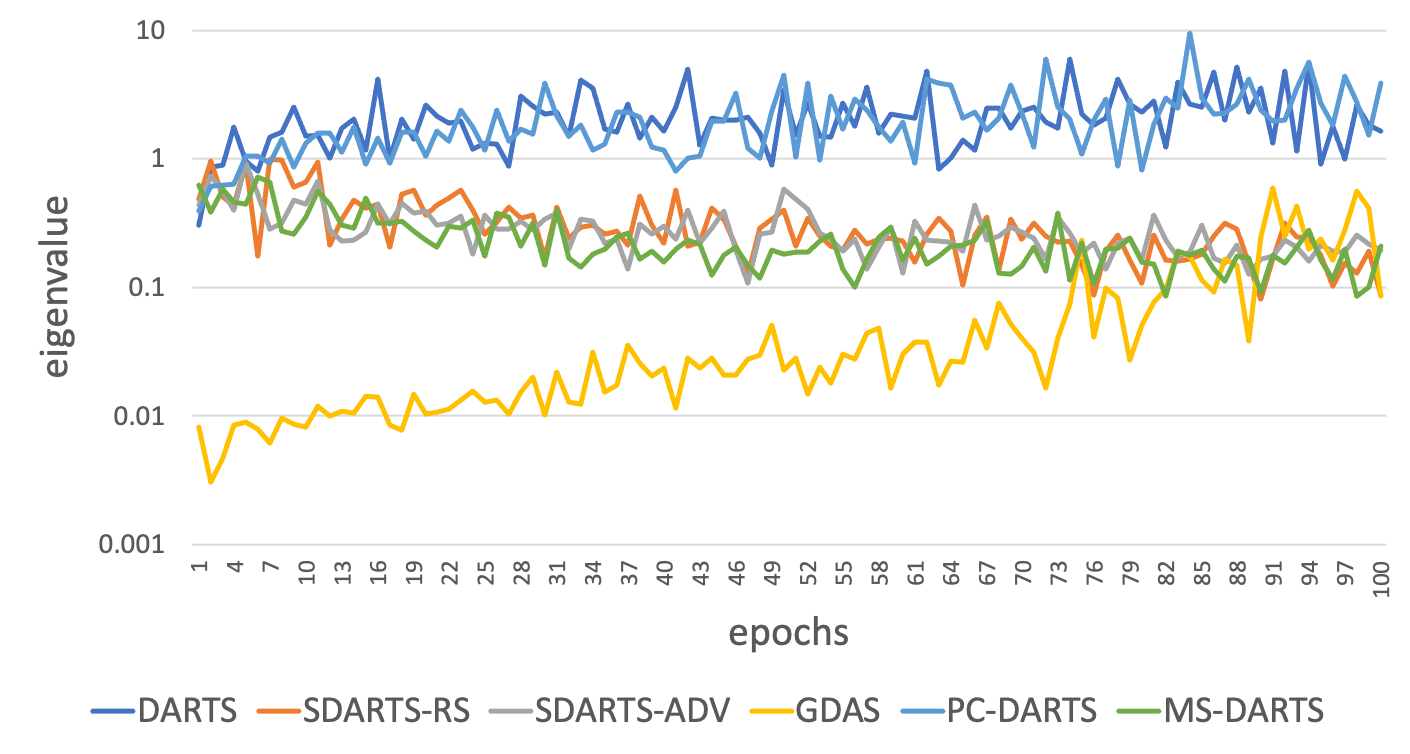}
    }
    \centerline{
        {\footnotesize (c) NAS-bench-1shot1 search space 3}
    }  
    \vspace{0.2cm}
    \caption{ Comparison of trajectories of the largest eigenvalues of Hessian matrices for architecture search on NAS-bench-1shot1.
    }
    \label{fig:evonnasbench}
\end{figure}

This dataset \cite{zela2020nasbench1shot1} consists of 3 search spaces based on CIFAR-10, with a provided mapping between the continuous space of differentiable NAS to the discrete space. Details are described in \cite{zela2019understanding,chen2020stabilizing}. Figure.\ref{fig:evonnasbench} shows the comparison of our models against five SoTA methods, namely DARTS \cite{DARTS:ICLR2019}, PC-DARTS \cite{PCDARTS:ICLR2020}, GDAS \cite{GDAS:CVPR2019}, and SDARTS-RS/ADV \cite{chen2020stabilizing} on all 3 search spaces.
We run every NAS algorithm for 100 epochs to allow thorough and comprehensive analysis on search stability and generalizability. All performance comparisons were evaluated on a V100 GPU. PC-DARTS outperforms the original DARTS. GDAS, SDARTS-RS, and SDARTS-ADV outperform PC-DARTS. However, GDAS suffers from a pre-mature convergence to sub-optimal architectures. Our MS-DARTS outperform all five comparison models. As shown in Figure.\ref{fig:evonnasbench}. The final eigenvalue of MS-DARTS is lower than all others, showing its capability in searching for DARTS searching stability.

%--------------------------------------
%--------------------------------------

%--------------------------------------
\subsection{Experiments of Wider Minima}
\label{sec:EXPwiderminima}
\begin{figure*}[t]
\centerline{
\includegraphics[height=0.3\columnwidth]{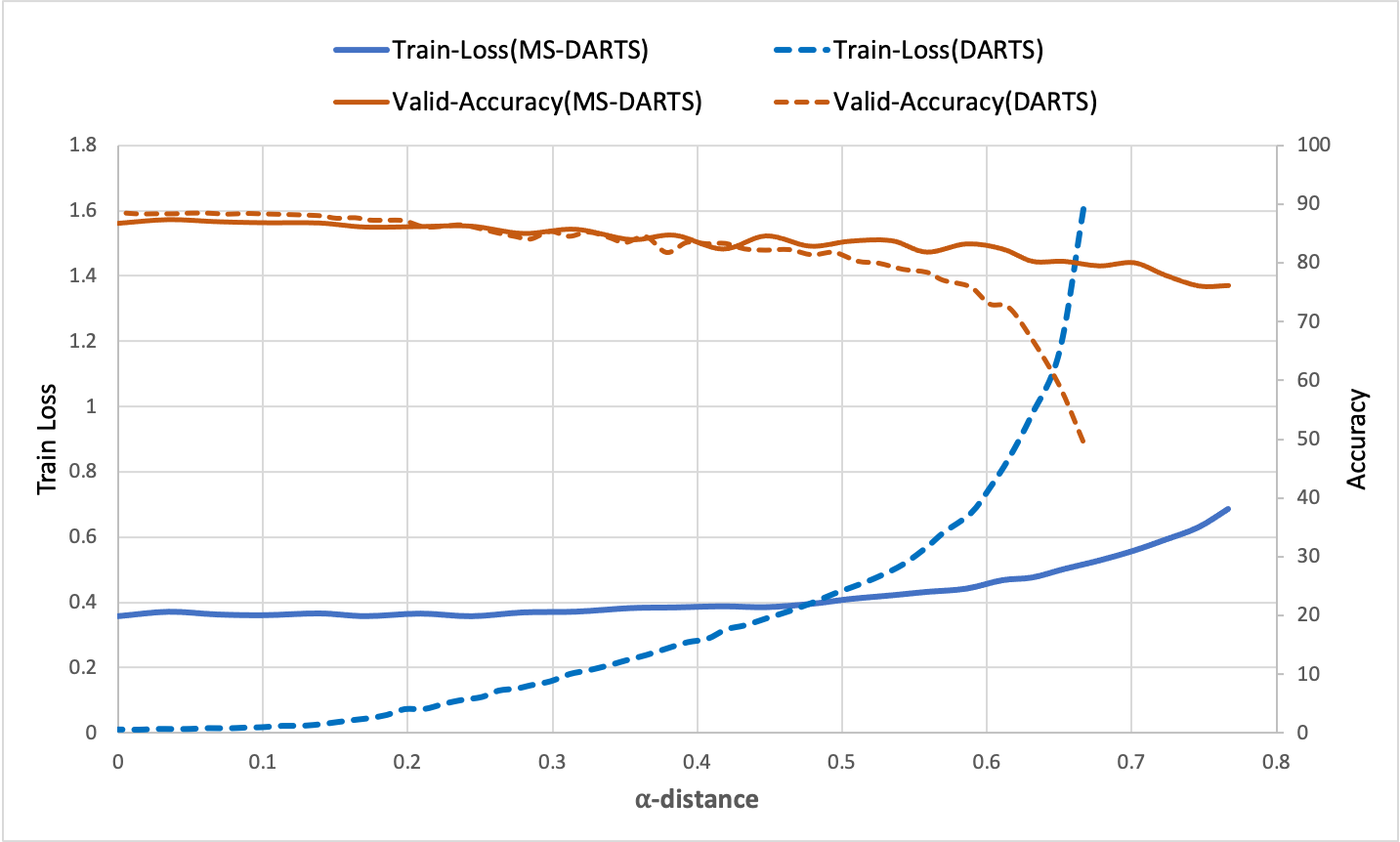}
\vspace{-0.1cm}
}
\caption{
$\alpha$-distance during training search on Cifar-10.
\vspace{-0.3cm}
}
\label{fig:alpha-distance}
\end{figure*}

Let $\alpha_t$ denote the {\em t}-th architecture obtained at the {\em t}-th epoch and $\alpha_{opt}$ be the final architecture after optimization.
Then, we can define the $\alpha$-distance  for $\alpha_t$ as $||\alpha_t-\alpha_{opt}||$. To prove our method can get wider minima, the training-loss vs. the $\alpha$-distance on CIFAR-10 was plotted in Figure.\ref{fig:alpha-distance}.
Clearly, the shape of train loss vs. $\alpha$–distance curve is considerably wider for our MS-DARTS than for the original DARTS, suggesting that MS-DARTS indeed converges to a wider solution. It also proves the stability of our MS-DARTS is better than the original DARTS.
Although the original DARTS gets lower training errors than our MS-DARTS, its sharp minima lead to higher evaluation testing errors than our MS-DARTS. It is noticed that the evaluation testing error for MS-DARTS is $2.51$ which is lower than the one for the original DARTS on CIFAR-10, {\em i.e.}, $2.76$ (see Table 1). Furthermore, we find that our MS-DARTS converges faster than previous DARTS works during experiments in Figure.\ref{fig:alpha-distance} since wider minima works. We further reduce the searching epochs from 50 to 40, this change gets lower training search time. The validation accuracy curves also prove the same observation and conclusion. The validation accuracy curve of MS-DARTS is flatter than the one of the original DARTS. A flatter accuracy surface makes our MS-DARTS stable to get higher test-accuracy although its validation accuracy is lower than DARTS.
% --------------------------------------------
% --------------------------------------------

\subsection{Experiments on Cifar10 dataset}
\label{sec:EXPcifar10}
\begin{figure}[b]
    \centerline{
        {\footnotesize (a)}
        \includegraphics[height=0.3\columnwidth]{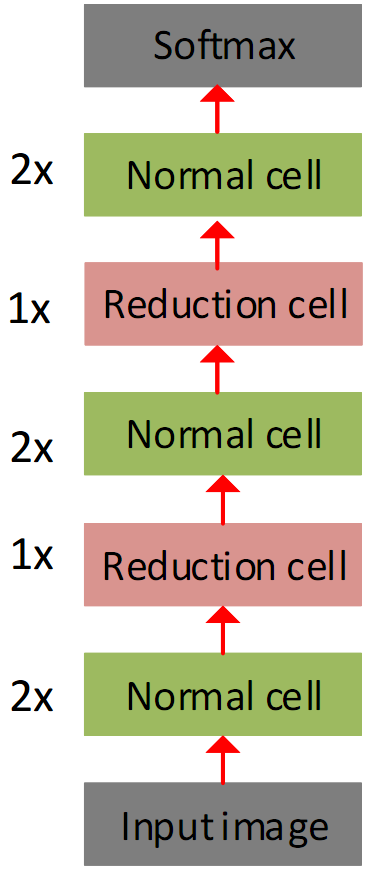}
        {\footnotesize (b)}
        \includegraphics[height=0.3\columnwidth]{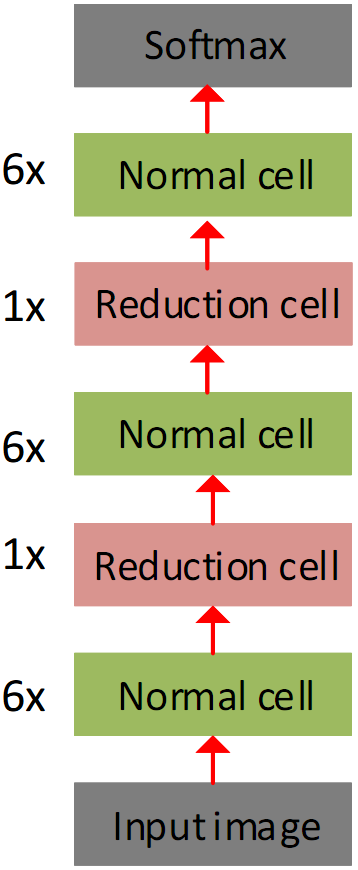}
    }
    \caption{
    Training search and evaluation stacked architecture
    (a) The MS-DARTS searched architecture on CIFAR-10.
    (b) MS-DARTS generated architecture used for evaluation on CIFAR-10.}
    \label{fig:TAEA}
\end{figure}

{\bf Training on CIFAR-10.}
We apply MS-DARTS to determine the convolutional cells on CIFAR-10~\cite{CIFAR-10} and then enlarges the network by stacking the learned cells. Figure.\ref{fig:TAEA}(a) shows the learned cell with $7$ ordered nodes ($2$ input nodes, $4$ intermediate nodes, and $1$ output node); $8$ of such learned cells are stacked to build the enlarged network. Each of these nodes is connected to the previous nodes with forecasted operations. The learned cell is also connected with the output from two previous cells.  The continuous searching architecture following in DARTS \cite{DARTS:ICLR2019} is including the following operations: {\em max\_pooling\_3$\times$3}, {\em avg\_pooling\_3$\times$3},
{\em skip\_connection}, {\em sep\_conv\_$3\times$3}, {\em sep\_conv\_$5\times$5}, {\em dil\_conv\_$3\times$3}, {\em dil\_conv\_5$\times$5}.
Different from \cite{yao2020efficient}, our results show that both the normal and reduction cells consist of combination of operations that are automatically selected from the search spaces. We set $40$ epochs with batch size $64$ on the CIFAR-10 dataset, which was equally split into training and validation set. We adopt the same weight optimization as in \cite{DARTS:ICLR2019} using the SGD optimizer with momentum $0.9$, weight decay $3\times{10}^{-4}$, and learning rate annealed from $0.025$ to ${10}^{-3}$. We found that $T=2,\ 3$ performs the best out of experiments of $T={2,\ 3,\ 4,\ 5}$, and $N={2,\ 3,\ 4}$ for different search spaces and datasets. For search space s5, every epoch takes nearly about 6 minutes on average on a single GPU. Architecture search training takes about 11 hrs.

\begin{figure}[t]
\centering
    \centerline{
        {\footnotesize (a)}
        \includegraphics[height=0.23\linewidth]{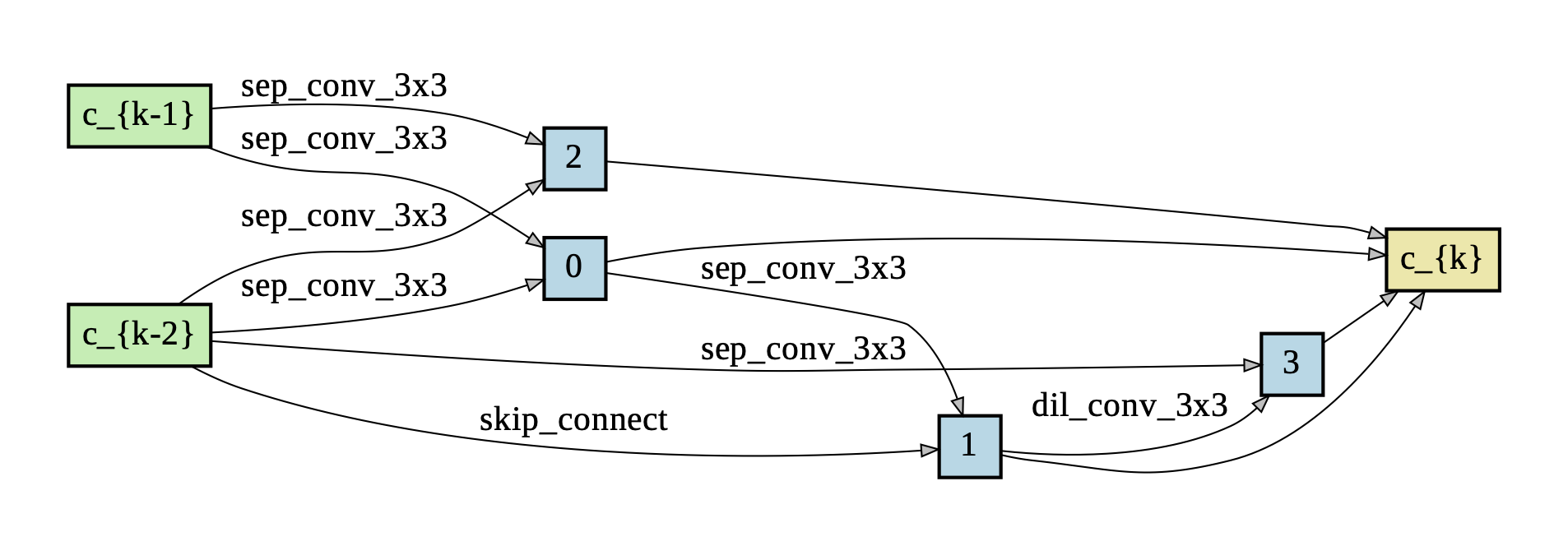}
    }
    \centerline{
        {\footnotesize (b)}
        \includegraphics[height=0.15\linewidth]{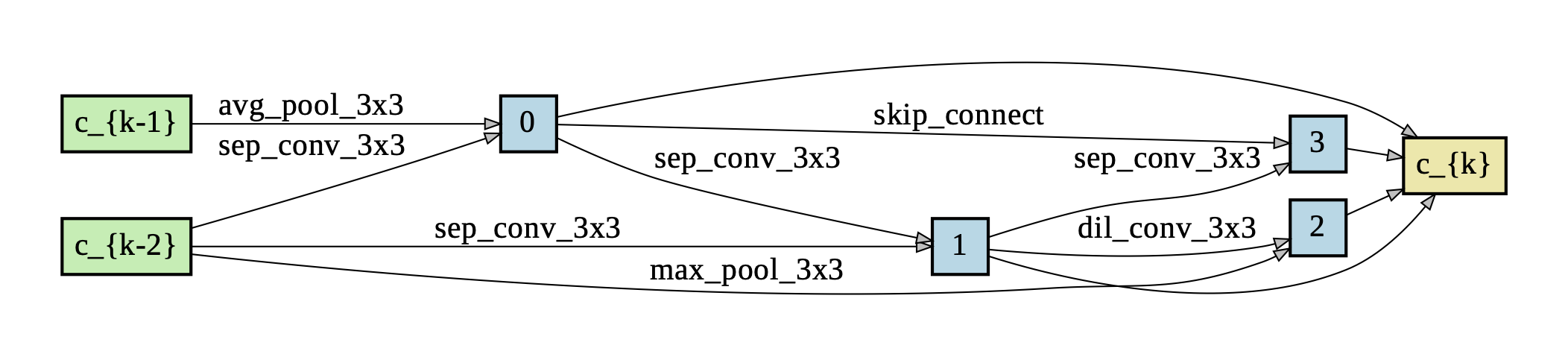}
    }
    \caption{
    MS-DARTS Normal and Reduce Cell.
    (a) MS-DARTS Normal Cell (b) MS-DARTS Reduction Cell.
    }
    \label{fig:NCRC}
    \vspace{-0.3cm}
\end{figure}

{\bf Evaluation on CIFAR-10.}
The parent architecture consists of $20$ learned cells ($18$ normal cells and $2$ reduction cells) and $36$ channels. Figure.\ref{fig:TAEA}(b) shows the evaluation architecture. Table~\ref{tab:CIFAR-10ER} shows the evaluation of $(mean\pm std)$ in $4$ independents runs with random seeds. The best architecture was selected based on accuracy. Compared with SDARTS-ADV\cite{chen2020stabilizing}, MS-DARTS only takes less than half of the training time with improved accuracy. Figure.\ref{fig:NCRC} shows the normal and reduction cells of MS-DARTS on CIFAR-10 dataset, where test error was $2.51\pm0.02$.

%--------------------------------------
\vspace{-0.1cm}
% --------------------------------------------
% --------------------------------------------
\begin{table*}
\footnotesize
\begin{center}
\begin{tabular}{lllll}
 %\hline
 %\multicolumn{5}{|c|}{Country List} \\
 \hline
 Architecture &
 \multicolumn{1}{|p{1.4cm}|}{Mean test error $(\%)$ } & 
 \multicolumn{1}{|p{1.2cm}|}{ Params (M)} & 
 \multicolumn{1}{|p{1.8cm}|}{ Search cost (GPU days)} &
 \multicolumn{1}{|p{2cm}}{Search method} \\
 \hline
 DenseNet-BC \cite{DenseNet:CVPR2017} & 3.46 & 25.6 & - & manual\\
 \hline
 NASNet-A \cite{NASNet:CVPR2018} & 2.65 & 3.3 & 2000 & RL \\
 ENAS \cite{ENAS:ICML2018} & 2.89 & 4.6 & 0.5 & RL \\
 NAS \cite{zoph2017neural} & 4.47 & 7.1 & 1800 & RL \\
 \hline
 AmoebaNet-A \cite{AmoebaNet:AAAI2019} & 3:34 $\pm 0.06$ & 3.2 & 3150 & evolution \\
 AmoebaNet-B \cite{AmoebaNet:AAAI2019} & 2:55 $\pm 0.05$ & 2.8 & 3150  & evolution \\
 Hierarchical Evolution \cite{Hier:Evol:ICLR2018} & 3.75 & 15.7 & 300 & evolution \\
 \hline
 PNAS \cite{Progressive:NAS:ECCV2018} & 3:41 $\pm 0.09$ & 3.2 & 225 & SMBO \\
 NAONet \cite{NAONet:NIPS2018} & 3.53 & 3.1 & 0.4 & NAO \\
 \hline
 SNAS (mild) +cutout \cite{xie2018snas}&2.98&2.9&1.5&gradient-based \\
 GDAS  + cutout \cite{GDAS:CVPR2019}&2.93&3.4&0.2&gradient-based \\
 DARTS (2nd) \cite{DARTS:ICLR2019}&2.76&3.4&0.3&gradient-based \\
 PC-DARTS  +cutout \cite{PCDARTS:ICLR2020}&2.57&3.6&0.1&gradient-based \\
 LA-DARTS  +cutout \cite{LADARTS:arXiv2020}&2.72&2.7&0.7 $^\dag$ & gradient-based \\
 SDARTS-RS \cite{chen2020stabilizing}&2.67 $\pm 0.03$&3.4&0.4 $^\dag$ & gradient-based\\
 SDARTS-ADV \cite{chen2020stabilizing}&2.61 $\pm 0.02$&3.3&1.3 $^\dag$ & gradient-based\\
 \hline
 \textbf{MS-DARTS} & {2.51 $\pm 0.02$} & 3.8 & 0.4 $^\dag$ & gradient-based\\
 \hline
\end{tabular}
\end{center}
\vspace{-0.3cm}
\caption{Comparison with SoTA image classifiers on CIFAR-10.
 $^\dag$ Tested on a Tesla-V100 GPU.
 }
\label{tab:CIFAR-10ER}
\vspace{-0.1cm}
\end{table*}
% --------------------------------------------

%%%%%%%%%%%%%%%%%%%%%%%%%%%%%%%%%%%%%%%%%%%%%

\subsection{Experiments on parameter-free operations}
As chapter \ref{sec:2.3} mentioned, excessive parameter-free operations such as noise, skip-connection and null let the generated architecture’s performance crash, though these operations in mixed-up continuous architecture could make architecture converge faster in training search process. R-DARTS \cite{zela2019understanding} proposed 4 simplified search spaces which only contain parameter-free and a portion of candidate operations to test searching algorithm’s regularizations. We implement our MS-DARTS and previous DARTS works across 2 datasets (Cifar-10/Cifar-100) on these 4 search spaces (S1-S4).
% --------------------------------------------
{\bf Training search on CIFAR-10 / Cifar-100.} As same setting in chapter \ref{sec:EXPcifar10}, the generated normal cell and reduction cell by our MS-DARTS searching algorithm will be transferred to Figure.\ref{fig:TAEA}(b) architecture. The results are shown in Table~\ref{tab:CIFAR-100_S1-4}.

{\bf Evaluation on CIFAR-100.} Contrary to Cifar-10, Cifar-100 has 100 classes and more training/testing images. We take the generated architecture (learned cells) created with CIFAR-100 and evaluate on CIFAR-100 \cite{cifar100}. To fairly compare with other DARTS algorithms, we using the same training setting as previous works which is different from chapter \ref{sec:EXPwiderminima}, the parent architecture consists of $8$ learned cells and $16$ channels. Channel numbers are doubled after each reduction cells. The parent architecture was trained by $600$ epochs using batch size $64$ and SGD optimizer with momentum $0.9$ and learning rate cosine scheduled from $0.025$ to 0. For regularization, schedule drop-path was used with $0$ to $0.2$ linear increase, auxiliary towers with weight $0.4$, and cutout \cite{Cutout:arXiv2017} data augmentation. In this experiment, all other parameters and optimization functions are kept intact. The training and validation tasks were performed using CIFAR-100 dataset with $4$ search spaces (see R-DARTS \cite{zela2019understanding}). Table~~\ref{tab:CIFAR-100_S1-4}  shows the comparisons of MS-DARTS methods against other SoTA methods transferred from CIFAR-10 training to CIFAR-100 evaluation in the $4$ search spaces.  Clearly, SDARTS-ADV outperforms DARTS, R-DARTS, DARTS-ES, and PC-DARTS. Our MS-DARTS also outperforms all SoTA DARTS methods with lower or similar variations.

%--------------------------------------------
\begin{table}[t]
\vspace{-0.3cm}
\centerline{
\setlength\tabcolsep{1.8pt}
\footnotesize
\begin{tabular}{lllrlllll}
\toprule
\multicolumn{1}{c}{Dataset} & \multicolumn{4}{c}{CIFAR-10}                     & \multicolumn{4}{c}{CIFAR-100}      \\
Method / Space              & S1   & S2   & \multicolumn{1}{l}{S3} & S4   & S1    & S2    & S3    & S4    \\
\toprule
DARTS        & 3.84 & 4.85 & 3.34 & 7.20 & 29.46 & 26.05 & 28.90 & 22.85 \\
PC-DARTS     & 3.11 & 3.02 & 2.51 & 3.02 & 24.69 & 22.48 & 21.69 & 21.50 \\
DARTS-ES     & 3.01 & 3.26 & 2.74 & 3.71 & 28.37 & 23.25 & 23.73 & 21.26 \\
R-DARTS (DP) & 3.11 & 3.48 & 2.93 & 3.58 & 25.93 & 22.30 & 22.36 & 22.18 \\
R-DARTS(L2)  & 2.78 & 3.31 & 2.51 & 3.56 & 24.25 & 22.44 & 23.99 & 21.94 \\
SDATRS-RS    & 2.78 & 2.75 & 2.53 & 2.93 & 23.51 & 22.28 & 21.09 & 21.46 \\
SDARTS-ADV   & 2.73 & 2.65 & 2.49 & 2.87 & 22.23 & 20.56 & 21.08 & 21.25 \\\hline
\textbf{MS-DARTS}   &2.70 &2.47  &2.46 &2.85 & 21.14 & 20.55 & 20.45 & 20.79 \\
\bottomrule
\end{tabular}
}
\caption{
Comparison with SoTA DARTS-based methods on CIFAR-10 and CIFAR-100 datasets.
}
\label{tab:CIFAR-100_S1-4}
\vspace{-0.1cm}
\end{table}

%-------------------------
\subsection{Experiments on ImageNet dataset}
{ \bf Evaluation on ImageNet.} ImageNet Classification \cite{ImageNet:CVPR2009} is a high-resolution image classification dataset proposed by Alex.{\em et.} This dataset consists of 1000 different classes and more 1.2 million images. Although ImageNet has been released from 2012, it still be the most famous dataset in image classification tasks. Not only in classification works, many object detection and image segmentation tasks also use the dataset for training convolution backbone. In view of this, we next compare MS-DARTS against SoTA methods on ImageNet. Similar to previous works R-DARTS \cite{zela2019understanding} and S-DARTS \cite{chen2020stabilizing}, the network we constructed consists of $14$ cells and $48$ channels. Parameters were trained for $300$ epochs using SGD optimizer with an annealing learning rate initialized as $0.5$, momentum $0.9$, and weight decay $3\times{10}^{-5}$. Table~\ref{tab:ImageNetER} compares 11 baseline methods on ImageNet evaluation. Both SDARTS-RS and SDARTS-ADV perform better than other DARTS by a large margin. MS-DARTS again outperforms other SoTA DARTS-based methods. Note that additional regularization techniques including partial channel connection \cite{PCDARTS:ICLR2020} can further improve the accuracy \cite{chen2020stabilizing}. For fair comparisons, those regularization techniques \cite{Progressive:Diff:NAS:ICCV2019} were not included here.

% Table ImagNet results
% --------------------------------------------
\begin{table}[t]
\vspace{-0.3cm}
\footnotesize
\begin{center}
\begin{tabular}{lllll}
\hline
Architecture & \multicolumn{2}{c}{Test Error $(\%)$} \\ & top-1 & top-5\\
 \hline
 Inception-v1 \cite{InceptionV1:CVPR2015} & 30.1 & 10.1 \\
 MobileNet \cite{MobileNets:arXiv2017} & 29.4 & 10.5 \\
 ShuffleNet-v2 \cite{ShuffleNetV2:ECCV2018} & 25.1 & 10.1 \\
 \hline
 NASNet-A \cite{NASNet:CVPR2018} & 26.0 & 8.4 \\
 PNAS \cite{Progressive:NAS:ECCV2018} & 25.8 & 8.1 \\
 AtomNAS-A \cite{AmoebaNet:AAAI2019} & 25.4 & 7.9 \\
 MnasNet-92 \cite{MnasNet:CVPR2019} & 25.2 & 8.0 \\
 
 \hline
 DARTS \cite{DARTS:ICLR2019} & 26.7 & 8.7 \\
 SNAS \cite{xie2018snas} & 27.3 & 9.2 \\
 SDARTS-RS \cite{chen2020stabilizing} & 25.6 & 8.5 \\
 SDARTS-ADV \cite{chen2020stabilizing} & 25.2 & 7.8 \\
 \hline
 {MS-DARTS} & \textbf {24.4} & \textbf{7.3} \\
 \hline
\end{tabular}
\end{center}
%\vspace{-0.5cm}
\caption{Comparison with SoTA image classifiers on ImageNet.}
\label{tab:ImageNetER}
\vspace{-0.5cm}
\end{table}
% --------------------------------------------
%%%%%%%%%%%%%%%%%%%%%%%%%%%%%%%%%%%%%%%%%%%%%%%%%
%%%%%%%%%%%%%%%%%%%%%%%%%%%%%%%%%%%%%%%%%%%%%%%%%
\subsection{Experiments of Bandwidth Effect w.r.t MS-DARTS Stability}
To verify our suppose in chapter \ref{sec:3.4}, we setting several numbers of the hyperparameter – bandwidth (h). At the same setting of other hyperparameter, likes sampling radius epsilon ($\epsilon$), mean-shift iteration ($T$), number of sampling points ($N$) and training search epoch ($E$), we test the effect of MS-DARTS stability with different bandwidth values on NAS-bench-1shot1 dataset. Table 4 shows the eigenvalue (mean$\pm$std) of each bandwidth value ($h$) during training search. Fitting our proposal in above-mentioned, a reasonably large bandwidth h that produces the best desired smoothed landscape and lower eigenvalue, but too large bandwidth h can over-smooth the landscape.

\begin{table}[t]
\vspace{-0.3cm}
\footnotesize
\begin{center}
\begin{tabular}{llllll}
\hline
{bandwidth} & \multicolumn{4}{c}{Epoch} & {Test} \\ 
\multicolumn{1}{c}{($h$)} & \multicolumn{1}{c}{1$\sim$20} & \multicolumn{1}{c}{21$\sim$40} & \multicolumn{1}{c}{41$\sim$60} & \multicolumn{1}{c}{61$\sim$80} & \multicolumn{1}{c}{error}  \\
\hline
0.2  & 0.59$\pm$0.25  & 0.33$\pm$0.12  & 0.22$\pm$0.10  & 0.18$\pm$0.08  & 0.067  \\
\hline
0.4  & 0.58$\pm$0.19  & 0.31$\pm$0.11  & 0.23$\pm$0.09  & 0.19$\pm$0.08  & 0.067  \\
\hline
0.6  & 0.59$\pm$0.20  & 0.30$\pm$0.12  & 0.24$\pm$0.07  & 0.16$\pm$0.07  & 0.067   \\
\hline
0.8  & 0.49$\pm$0.13  & 0.32$\pm$0.12  & 0.22$\pm$0.05  & \textbf{0.15}$\pm$0.06  & 0.064   \\
\hline
1.0  & 0.58$\pm$0.28  & 0.30$\pm$0.12  & 0.23$\pm$0.09  & \textbf{0.15}$\pm$\textbf{0.04}  & \textbf{0.061}  \\
\hline
1.2  & 0.52$\pm$0.18  & 0.30$\pm$0.09  & 0.24$\pm$0.08  & 0.17$\pm$0.09  & 0.064    \\
\hline
1.4  & 0.53$\pm$0.19  & 0.36$\pm$0.11  & 0.23$\pm$0.08  & 0.17$\pm$0.06  & 0.067     \\
\hline
\end{tabular}
\end{center}
%\vspace{-0.5cm}
\caption{ Comparison eigenvalues with several bandwidth value on NASbench-1shot1 s1.}
\label{tab:Bandwidth}
\vspace{-0.5cm}
\end{table}

%%%%%%%%%%%%%%%%%%%%%%%%%%%%%%%%%%%%%%%%%%%%%%%%%
%%%%%%%%%%%%%%%%%%%%%%%%%%%%%%%%%%%%%%%%%%%%%%%%%
\vspace{-0.1cm}
\section{Conclusion}
We presented a mean-shift based DARTS approach to stabilize and improve DARTS network architecture search. Experimental results also prove that the mean-shift design can smooth out the sharp minima caused by DARTS-based NAS to flatter minima from which significant performance drop can be avoided. We also investigated the selection of the bandwidth hyper-parameter and the convergence of the proposed mean-shift filtering. Extensive experiments on CIFAR-10, CIFAR-100, and ImageNet show that our method outperforms various state-of-the-art DART methods. The proposed DARTS stabilization is advantageous and general. MS-DARTS is useful for practical applicability that it can work on high-resolution images provided sufficient GPU computation. 

{\bf Future work.} MS-DARTS can be extended to explore additional network components in other application domains, to generate models with better accuracy in reduced time.

\clearpage

\bibliographystyle{unsrt}  
\bibliography{references}  

\begin{thebibliography}{10}

\bibitem{NAS:Survey:JMLR2019}
Thomas Elsken, Jan~Hendrik Metzen, and Frank Hutter.
\newblock Neural architecture search: A survey.
\newblock {\em JMLR}, pages 1--21, 2019.

\bibitem{zoph2017neural}
Barret Zoph and Quoc~V Le.
\newblock Neural architecture search with reinforcement learning.
\newblock {\em ICLR}, 2017.

\bibitem{DARTS:ICLR2019}
Hanxiao Liu, Karen Simonyan, and Yiming Yang.
\newblock {DARTS}: Differentiable architecture search.
\newblock In {\em ICLR}, 2019.

\bibitem{AmoebaNet:AAAI2019}
Esteban Real, Alok Aggarwal, Yanping Huang, and Quoc~V Le.
\newblock Regularized evolution for image classifier architecture search.
\newblock In {\em AAAI}, volume~33, pages 4780--4789, 2019.

\bibitem{Hier:Evol:ICLR2018}
Hanxiao Liu, Karen Simonyan, Oriol Vinyals, Chrisantha Fernando, and Koray
  Kavukcuoglu.
\newblock Hierarchical representations for efficient architecture search.
\newblock In {\em ICLR}, 2018.

\bibitem{NASNet:CVPR2018}
Barret Zoph, Vijay Vasudevan, Jonathon Shlens, and Quoc~V Le.
\newblock Learning transferable architectures for scalable image recognition.
\newblock In {\em CVPR}, pages 8697--8710, 2018.

\bibitem{ENAS:ICML2018}
Hieu Pham, Melody Guan, Barret Zoph, Quoc Le, and Jeff Dean.
\newblock Efficient neural architecture search via parameters sharing.
\newblock In {\em ICML}, pages 4095--4104, 2018.

\bibitem{PCDARTS:ICLR2020}
Yuhui Xu, Lingxi Xie, Xiaopeng Zhang, Xin Chen, Guo-Jun Qi, Qi~Tian, and
  Hongkai Xiong.
\newblock {PC-DARTS}: Partial channel connections for memory-efficient
  architecture search.
\newblock In {\em ICLR}, 2020.

\bibitem{xie2018snas}
Sirui Xie, Hehui Zheng, Chunxiao Liu, and Liang Lin.
\newblock {SNAS}: stochastic neural architecture search.
\newblock In {\em ICLR}, 2019.

\bibitem{LADARTS:arXiv2020}
Yuhui Xu, Lingxi Xie, Xiaopeng Zhang, Xin Chen, Bowen Shi, Qi~Tian, and Hongkai
  Xiong.
\newblock Latency-aware differentiable neural architecture search.
\newblock {\em arXiv:2001.06392}, 2020.

\bibitem{zela2019understanding}
Arber Zela, Thomas Elsken, Tonmoy Saikia, Yassine Marrakchi, Thomas Brox, and
  Frank Hutter.
\newblock Understanding and robustifying differentiable architecture search.
\newblock In {\em ICLR}, 2020.

\bibitem{Eval:Search:Phase:NAS:ICLR2020}
Kaicheng Yu, Christian Sciuto, Martin Jaggi, Claudiu Musat, and Mathieu
  Salzmann.
\newblock Evaluating the search phase of neural architecture search.
\newblock {\em ICLR}, 2020.

\bibitem{Cell:NAS:ICLR2019}
Yao Shu, Wei Wang, and Shaofeng Cai.
\newblock Understanding architectures learnt by cell-based neural architecture
  search.
\newblock In {\em ICLR}, 2019.

\bibitem{DAAS:arXiv2020}
Yunjie Tian, Chang Liu, Lingxi Xie, Jianbin Jiao, and Qixiang Ye.
\newblock Discretization-aware architecture search.
\newblock {\em arXiv:2007.03154}, 2020.

\bibitem{CIFAR-10}
Alex Krizhevsky, Vinod Nair, and Geoffrey Hinton.
\newblock {CIFAR-10} dataset.
\newblock \url{http://www.cs.toronto.edu/~kriz/cifar.html}.

\bibitem{MeanShift:PAMI2002}
Dorin Comaniciu and Peter Meer.
\newblock Mean shift: A robust approach toward feature space analysis.
\newblock {\em IEEE Trans. PAMI}, 24(5):603--619, 2002.

\bibitem{NAS:RL:ICLR2017}
Bowen Baker, Otkrist Gupta, Nikhil Naik, and Ramesh Raskar.
\newblock Designing neural network architectures using reinforcement learning.
\newblock {\em ICLR}, 2017.

\bibitem{Block:DNA:CVPR2020}
Changlin Li, Jiefeng Peng, Liuchun Yuan, Guangrun Wang, Xiaodan Liang, Liang
  Lin, and Xiaojun Chang.
\newblock Block-wisely supervised neural architecture search with knowledge
  distillation.
\newblock In {\em CVPR}, pages 1989--1998, 2020.

\bibitem{DAS:CVPR2018}
Richard Shin, Charles Packer, and Dawn Song.
\newblock Differentiable neural network architecture search.
\newblock In {\em CVPR}, 2018.

\bibitem{Maskconnect:ECCV2018}
Karim Ahmed and Lorenzo Torresani.
\newblock {Maskconnect}: Connectivity learning by gradient descent.
\newblock In {\em ECCV}, pages 349--365, 2018.

\bibitem{ResNet:CVPR2016}
Kaiming He, Xiangyu Zhang, Shaoqing Ren, and Jian Sun.
\newblock Deep residual learning for image recognition.
\newblock In {\em CVPR}, pages 770--778, 2016.

\bibitem{DenseNet:CVPR2017}
Gao Huang, Zhuang Liu, Laurens Van Der~Maaten, and Kilian~Q Weinberger.
\newblock Densely connected convolutional networks.
\newblock In {\em CVPR}, pages 4700--4708, 2017.

\bibitem{CNF:NIPS2016}
Shreyas Saxena and Jakob Verbeek.
\newblock Convolutional neural fabrics.
\newblock In {\em NeurIPS}, pages 4053--4061, 2016.

\bibitem{Budgeted:SuperNet:CVPR2018}
Tom Veniat and Ludovic Denoyer.
\newblock Learning time/memory-efficient deep architectures with budgeted super
  networks.
\newblock In {\em CVPR}, pages 3492--3500, 2018.

\bibitem{Progressive:Diff:NAS:ICCV2019}
Xin Chen, Lingxi Xie, Jun Wu, and Qi~Tian.
\newblock Progressive differentiable architecture search: Bridging the depth
  gap between search and evaluation.
\newblock In {\em ICCV}, pages 1294--1303, 2019.

\bibitem{XNAS:NIPS2019}
Niv Nayman, Asaf Noy, Tal Ridnik, Itamar Friedman, Rong Jin, and Lihi Zelnik.
\newblock {XNAS}: Neural architecture search with expert advice.
\newblock In {\em NeurIPS}, pages 1977--1987, 2019.

\bibitem{DARTS+:arXiv2019}
Hanwen Liang, Shifeng Zhang, Jiacheng Sun, Xingqiu He, Weiran Huang, Kechen
  Zhuang, and Zhenguo Li.
\newblock {DARTS+}: Improved differentiable architecture search with early
  stopping.
\newblock {\em arXiv:1909.06035}, 2019.

\bibitem{SGAS:CVPR2020}
Guohao Li, Guocheng Qian, Itzel~C Delgadillo, Matthias Muller, Ali Thabet, and
  Bernard Ghanem.
\newblock {SGAS}: Sequential greedy architecture search.
\newblock In {\em CVPR}, pages 1620--1630, 2020.

\bibitem{FairDARTS:ECCV2020}
Xiangxiang Chu, Tianbao Zhou, Bo~Zhang, and Jixiang Li.
\newblock Fair {DARTS}: Eliminating unfair advantages in differentiable
  architecture search.
\newblock In {\em ECCV}, 2020.

\bibitem{Random:UAI2020}
Liam Li and Ameet Talwalkar.
\newblock Random search and reproducibility for neural architecture search.
\newblock In {\em Uncertainty in Artificial Intelligence}, pages 367--377.
  PMLR, 2020.

\bibitem{MnasNet:CVPR2019}
Mingxing Tan, Bo~Chen, Ruoming Pang, Vijay Vasudevan, Mark Sandler, Andrew
  Howard, and Quoc~V Le.
\newblock Mnas{N}et: Platform-aware neural architecture search for mobile.
\newblock In {\em CVPR}, pages 2820--2828, 2019.

\bibitem{chen2020stabilizing}
Xiangning Chen and Cho-Jui Hsieh.
\newblock Stabilizing differentiable architecture search via perturbation-based
  regularization.
\newblock In {\em ICML}, 2020.

\bibitem{GradDensity:TIT1975}
Keinosuke Fukunaga and Larry Hostetler.
\newblock The estimation of the gradient of a density function, with
  applications in pattern recognition.
\newblock {\em IEEE Transactions on Information Theory}, 21(1):32--40, 1975.

\bibitem{MeanShift:PAMI1995}
Yizong Cheng.
\newblock Mean shift, mode seeking, and clustering.
\newblock {\em IEEE Trans. PAMI}, 17(8):790--799, 1995.

\bibitem{silverman1986density}
Bernard~W Silverman.
\newblock {\em Density estimation for statistics and data analysis}, volume~26.
\newblock CRC press, 1986.

\bibitem{Jastrzebski2018FindingFM}
Stanislaw Jastrzebski, Zachary Kenton, D.~Arpit, Nicolas Ballas, Asja Fischer,
  Yoshua Bengio, and A.~Storkey.
\newblock Finding flatter minima with sgd.
\newblock In {\em ICLR}, 2018.

\bibitem{FlatMinima}
Sepp Hochreiter and Jurgen Schmidhuber.
\newblock Flat minima.
\newblock {\em Neural computation}, 9:1--42, 02 1997.

\bibitem{AWL}
Pavel Izmailov, Dmitrii Podoprikhin, Timur Garipov, Dmitry~P. Vetrov, and
  Andrew~Gordon Wilson.
\newblock Averaging weights leads to wider optima and better generalization.
\newblock {\em CoRR}, abs/1803.05407, 2018.

\bibitem{JeanCMB14}
S{\'{e}}bastien Jean, Kyunghyun Cho, Roland Memisevic, and Yoshua Bengio.
\newblock On using very large target vocabulary for neural machine translation.
\newblock {\em CoRR}, abs/1412.2007, 2014.

\bibitem{ALAS}
Yakov~Z. Tsypkin and S.~J. Nikolic.
\newblock {\em Adaptation and Learning in Automatic Systems}.
\newblock Academic Press, Inc., USA, 1971.

\bibitem{zela2020nasbench1shot1}
Arber Zela, Julien Siems, and Frank Hutter.
\newblock Nas-bench-1shot1: Benchmarking and dissecting one-shot neural
  architecture search.
\newblock In {\em ICLR}, 2020.

\bibitem{cifar100}
Alex Krizhevsky, Vinod Nair, and Geoffrey Hinton.
\newblock Cifar-100 (canadian institute for advanced research).

\bibitem{ImageNet:CVPR2009}
J.~Deng, W.~Dong, R.~Socher, L.-J. Li, K.~Li, and L.~Fei-Fei.
\newblock {ImageNet}: A large-scale hierarchical image database.
\newblock In {\em CVPR}, 2009.

\bibitem{GDAS:CVPR2019}
Xuanyi Dong and Yi~Yang.
\newblock Searching for a robust neural architecture in four {GPU} hours.
\newblock In {\em CVPR}, pages 1761--1770, 2019.

\bibitem{yao2020efficient}
Quanming Yao, Ju~Xu, Wei-Wei Tu, and Zhanxing Zhu.
\newblock Efficient neural architecture search via proximal iterations.
\newblock In {\em AAAI}, pages 6664--6671, 2020.

\bibitem{Progressive:NAS:ECCV2018}
Chenxi Liu, Barret Zoph, Maxim Neumann, Jonathon Shlens, Wei Hua, Li-Jia Li,
  Li~Fei-Fei, Alan Yuille, Jonathan Huang, and Kevin Murphy.
\newblock Progressive neural architecture search.
\newblock In {\em ECCV}, pages 19--34, 2018.

\bibitem{NAONet:NIPS2018}
Renqian Luo, Fei Tian, Tao Qin, Enhong Chen, and Tie-Yan Liu.
\newblock Neural architecture optimization.
\newblock In {\em NeurIPS}, pages 7816--7827, 2018.

\bibitem{Cutout:arXiv2017}
Terrance DeVries and Graham~W Taylor.
\newblock Improved regularization of convolutional neural networks with cutout.
\newblock {\em arXiv:1708.04552}, 2017.

\bibitem{InceptionV1:CVPR2015}
Christian Szegedy, Wei Liu, Yangqing Jia, Pierre Sermanet, Scott Reed, Dragomir
  Anguelov, Dumitru Erhan, Vincent Vanhoucke, and Andrew Rabinovich.
\newblock Going deeper with convolutions.
\newblock In {\em CVPR}, pages 1--9, 2015.

\bibitem{MobileNets:arXiv2017}
Andrew~G Howard, Menglong Zhu, Bo~Chen, Dmitry Kalenichenko, Weijun Wang,
  Tobias Weyand, Marco Andreetto, and Hartwig Adam.
\newblock {MobileNets}: Efficient convolutional neural networks for mobile
  vision applications.
\newblock {\em arXiv:1704.04861}, 2017.

\bibitem{ShuffleNetV2:ECCV2018}
Ningning Ma, Xiangyu Zhang, Hai-Tao Zheng, and Jian Sun.
\newblock {ShuffleNet} v2: Practical guidelines for efficient {CNN}
  architecture design.
\newblock In {\em ECCV}, 2018.

\end{thebibliography}

\end{document}